\documentclass[letterpaper]{article}
\usepackage{aaai25} 
\usepackage{times} 
\usepackage{helvet} 
\usepackage{courier} 
\usepackage[hyphens]{url} 
\usepackage{graphicx} 
\usepackage{graphicx}  
\usepackage{natbib}  
\usepackage{caption}  
\usepackage{booktabs}
\usepackage{multirow}
\usepackage{hyperref}

\usepackage{xcolor}

\title{Script-Based Dialog Policy Planning for LLM-Powered Conversational Agents: \\ A Basic Architecture for an ``AI Therapist"}
\author{
   Robert Wasenmüller\textsuperscript{\rm 1},
   Kevin Hilbert\textsuperscript{\rm 2},
   Christoph Benzmüller\textsuperscript{\rm 3,\rm 4}
}
\affiliations {
   \textsuperscript{\rm 1}Aury, Berlin, Germany\\
   \textsuperscript{\rm 2}Department of Psychology, HMU Health and Medical University Erfurt, Erfurt, Germany,\\
   \textsuperscript{\rm 3}AI Systems Engineering, Otto-Friedrich Universität Bamberg, Bamberg, Germany\\
    \textsuperscript{\rm 4}Department of Mathematics and Computer Science, Freie Universität Berlin, Berlin, Germany\\
   robert.wasenmueller@aury.co,
   kevin.hilbert@hmu-erfurt.de,
   christoph.benzmueller@uni-bamberg.de
}
\date{August 2024}

\begin{document}
\maketitle
\vspace{-1em}
\begin{center}
    {\large August 2024}
\end{center}
\begin{abstract}

Large Language Model (LLM)-Powered Conversational Agents have the potential to provide users with scaled behavioral healthcare support, and potentially even deliver full-scale ``AI therapy'' in the future. While such agents can already conduct fluent and proactive emotional support conversations, they inherently lack the ability to (a) consistently and reliably act by predefined rules to align their conversation with an overarching therapeutic concept and (b) make their decision paths inspectable for risk management and clinical evaluation — both essential requirements for an ``AI Therapist''.

In this work, we introduce a novel paradigm for dialog policy planning in conversational agents enabling them to (a) act according to an expert-written ``script'' that outlines the therapeutic approach and (b) explicitly transition through a finite set of states over the course of the conversation. The script acts as a deterministic component, constraining the LLM's behavior in desirable ways and establishing a basic architecture for an AI Therapist.

We implement two variants of Script-Based Dialog Policy Planning using different prompting techniques and synthesize a total of 100 conversations with LLM-simulated patients. The results demonstrate the feasibility of this new technology and provide insights into the efficiency and effectiveness of different implementation variants.

\end{abstract}

\section{Introduction}

Large Language Model (LLM)-based conversational agents have gained in popularity in recent years, and have demonstrated exceptional proficiency in understanding context and generating responses across various domains \cite{Deng23a}.

Following their popularity, there is increased interest in academia and industry in using such agents in the behavioral health domain, including assisting therapists and conducting therapy. For instance, \citet{Stade24} describe the most advanced form of a clinical LLM as an autonomous psychotherapy AI, which we further call ``AI Therapist''. Such an LLM agent might feature all aspects of traditional therapy, including conducting comprehensive assessments, selecting appropriate interventions, and delivering a full course of therapy; all without oversight of human providers. This clearly has the potential to address insufficient capacity of mental healthcare systems around the world, and provide more individuals with access to personalized treatment.

However, psychotherapy is a highly complex and high-risk domain; interventions considerably affect patient's health, and require properly handling risk of harm to oneself and others. A full-fledged AI Therapist thus poses considerable ethical questions. With respect to the technical development, LLMs seem a well-suited foundation, but additional technologies may be necessary to ensure sufficient effectiveness and safety of such a system.

Our aim is to contribute to the development of an effective and safe LLM-based AI Therapist by proposing a basic LLM-based architecture that meets its general requirements.

In Section 2, we define five key requirements for an AI Therapist agent and derive technical methods well-suited to implement those requirements drawing from recent literature. We find that our requirements are fulfilled by \textit{LLM-Based Proactive Conversational Agents} complemented and constrained by a deterministic component inspired by handcrafted chatbots: A ``script'' outlining the conversational flow and instructions of the agent. We introduce our design of the script and the associated technique we call ``Script-Based Dialog Policy Planning''.

In Section 3, we propose two specific implementation variants for the AI Therapist powered by Script-Based Dialog Policy Planning based on an exemplary script and patient cases. We also suggest evaluation metrics suitable for comparing the efficiency and effectiveness of both variants.

We synthesize 100 dialogs between our AI Therapists and LLM-simulated patients, and outline the results in Section 4. We conclude the general feasibility of our novel approach and discuss the strengths and weaknesses of the design decisions in our implementation variants.

\section{Method}

\subsection{Requirements for an AI Therapist}

We argue that an AI Therapist needs to fulfill the following five requirements.

The first two are inspired by recent investigation on \textit{proactivity} of emotional support agents by \citet{Zheng23,Deng23a} and others.

\begin{itemize}
    \item[\textbf{(1)}] \textbf{Conversational Fluency}
Firstly, an AI Therapist needs to be able to hold a ``natural'' conversation, i.e. understand and remember context, as well as generate natural and high-quality responses, including reactions to various user utterances, even highly unexpected ones. We call this \textit{conversational fluency}.

\end{itemize}

\begin{itemize}
    \item[\textbf{(2)}] \textbf{Proactivity}
LLMs are originally trained to \textit{passively} follow users’ instructions, and prioritize accommodating users’ intentions \cite{Deng23b}.

We argue that an AI Therapist needs to act \textit{proactively}, i.e. be able to take the \textit{initiative} and control the conversation \cite{jurafsky24}.
\end{itemize}

The next three requirements are directly based on the guidelines for the responsible development of clinical LLMs proposed by \citet{Stade24}.

\begin{itemize}
    \item[\textbf{(3)}] \textbf{Expert Development}
Psychotherapists and behavioral health experts should be directly involved in the development of such agents, i.e. be able to explicitly define the agent’s functionality.

\item[\textbf{(4)}] \textbf{Application of Evidence-Based Practices}
The agent should apply evidence-based treatment practices (EBPs) only, with close adherence to its defined functionality and safeguarding practices for expectable risks. Beyond the specific interventions, this includes aligning the current conversation with the requirements of the overall therapeutic rationale, and embedding the current conversation in a larger context that includes a series of past and future conversations.

\item[\textbf{(5)}] \textbf{Inspectability}
Since inspectability and explainability of the agent’s behavior are clearly essential for risk management and clinical evaluation of such agents, we need an available trace through some pre-defined (finite) set of possible system states, including decisions made to move from one state to another.

\end{itemize}

\subsection{LLM-Based Proactive Conversational Agents}

\textit{LLM-Based Conversational Agents} are widely regarded as state-of-the-art technology to fulfill the requirement of \textit{(1) conversational fluency} \cite{Deng23b}. The ability to \textit{(2) proactively} steer the conversation towards specific goals, and thus tackle complicated tasks including strategic targets and motivational interactions requires LLM-Based \textit{Proactive} Conversational Agents (PCA) investigated in \cite{wu19}, \cite{Deng23a} and other recent works.

\subsubsection{Implementation Approaches}

There are two general approaches of implementing such agents \cite{cheng24}:

(1) \textit{Corpus-Based} learning approaches train the agent on a given set of data to predict the next dialog act. They rely on large datasets and/or human annotation, and are considered as inferior in optimizing for the long-term goal of the conversation. Such approaches are extended with planning techniques such as knowledge graphs \cite{yang22} and stochastic processes \cite{jianwang23}. However, they do not suffice for our requirement for explicit formulation and following of expert instructions.

(2) \textit{Prompt-Based} approaches prompt an LLM to conduct self-thinking for planning on each turn. They are based on explicit instructions and/or examples inside prompts. We further elaborate on such prompting techniques below.

\subsubsection{Proactive Dialog Policy Planning}

The agent’s underlying strategy for deciding on the next act is referred to as \textit{dialog policy} \cite{Harms19}. And the process of deciding what actions the dialog agent should take to achieve the specified goals during the conversation is called \textit{dialog policy planning} (DPP) \cite{Deng23b}.

\subsubsection{Proactive Prompting Techniques}

Multiple proactivity prompting techniques for LLM-Based PCAs have been proposed and investigated in recent works, for instance:

\begin{itemize}
\item (Basic) \textbf{Proactive Prompting} provides multiple options to the LLM to decide what act to take, and then generate a response based on the selected act \cite{Deng23a}.
\item \textbf{Proactive Chain-of-Thought} (ProCoT) is an extension of Proactive Prompting in which the LLM is instructed to perform one or multiple intermediate analysis steps as ``thoughts'' before deciding on one of the given dialog acts and generating a response \cite{Deng23Procot}.
\item \textbf{Ask-an-Expert} prompting scheme is involving another LLM actor, the ``expert'', which on each turn is given the context of the conversation and asked to answer one or multiple questions, helping the main LLM to generate its response to the user \cite{Zhang23}.
\end{itemize}

Further approaches are \textit{Cue-Based Chain-of-Thought} \cite{Wang23}, \textit{Tree-of-Thought} \cite{yao23}, \textit{Graph-of-Thought} \cite{besta23}, \textit{RePrompting} \cite{chen24}, and others. We can generalize that all proactive prompting approaches run through a similar set of steps on each turn of the conversation:

\begin{enumerate}
    \item (Optional) Reasoning and planning steps to prepare the decision and response generation
    \item Deciding on the next act from a given set of options
    \item Generating the response to the user
\end{enumerate}

\begin{figure}
    \centering
    \includegraphics[width=1\linewidth]{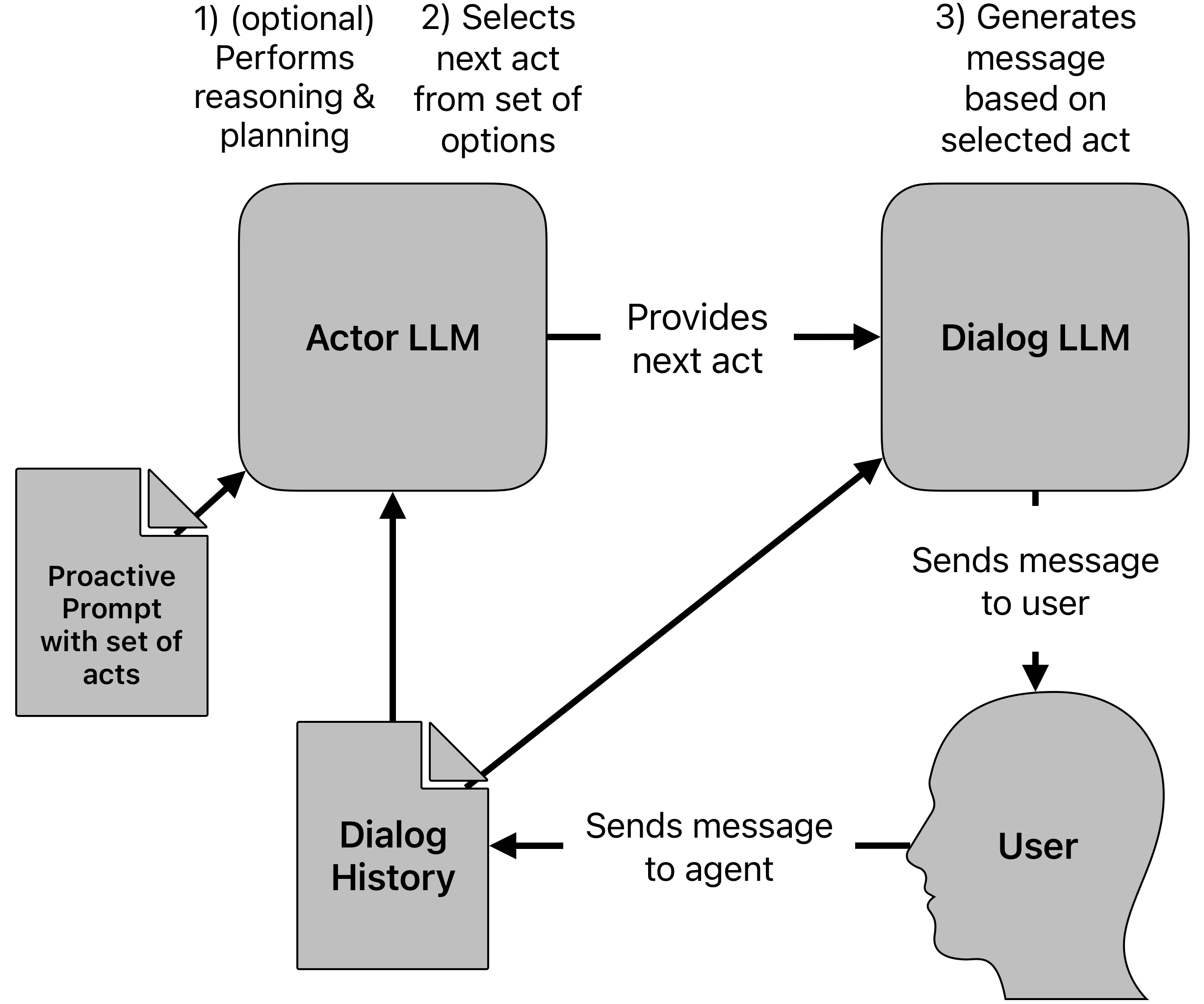}
    \caption{Proactive prompt-based dialog policy planning, version with an external Actor LLM: Steps on each conversation turn.}
    \label{fig:proactive-dpp}
\end{figure}

While step 3 is always performed by the main ``Dialog LLM'', steps 1 and 2 can either be performed by the Dialog LLM (like at ProCoT) or by an additional ``Actor LLM'' (like at Ask-an-Expert). See Figure \ref{fig:proactive-dpp} for an illustration of prompt-based proactive dialog policy planning \cite{Deng23b}.

\subsection{Script-Based Dialog Policy Planning}

The latter three requirements --- (3) expert development, (4) application of evidence-based practices, and (5) inspectability --- demand the following criteria:

\begin{itemize}
\item There needs to be a set of rules defined by experts,
\item the agent needs to reliably and constantly follow them,
\item the agent's decisions and acts must be explicitly and explainably selected from some finite set.
\end{itemize}

LLMs are primarily acting based on their training data and the user's immediate requests, i.e. not by explicit expert-defined rules. They also have infinite options for their decisions and acts, with no explainability of the decision process. However, the above criteria happen to be covered by traditional \textit{handcrafted / rule-based} dialog management approaches used to build chatbots that act deterministically upon a set of rules defined by their developers, and are modeled e.g. as finite-state machines \cite{Harms19}.

\subsubsection{Script for Policy Planning}

We introduce a component inspired by the \textit{handcrafted} dialog management approach: The ``script'', which is a natural language piece of text outlining two kinds of content:

\begin{itemize}
\item[\textbf{(1)}] \textbf{Conversational Flow} of the agent defining a finite set of \textit{states} for the agent as well as rules on when and how to transition from one state to another, and
\item[\textbf{(2)}] \textbf{State Instructions}, i.e. the instructions the agent should follow in each of the states.
\end{itemize}

Hence, we arrive at a \textit{hybrid dialog management} approach, which is known for combining advantages of both, the probabilistic and the rule-based approaches \cite{Harms19}. Compared to the hybrid DM approaches investigated by \citet{pande23} and \citet{kelley24}, we start with a data-driven LLM agent to ensure a high degree of conversational fluency, and add a rule-based component to constrain it.

The script can be written and iterated in natural language by domain experts with no background in software development. A JSON-formatted script like in our example below might be easier to interpret for the software, but is not necessary as LLMs can be asked to interpret unformatted text.

The concept of a script is supported by the fact that there are a large number of psychotherapeutic manuals, which in a sense serve as prompts for human therapists.

\subsubsection{Injecting Instructions via Prompts}

Due to the lack of existing therapy transcripts, implementing a \textit{corpus-based} approach would require a costly process of translating the script into human-written or synthesized dialogs, and training the LLM based on those. At the same time, in a corpus-based approach, it is more difficult to identify which state the agent is in, and check its adherence to the script. \cite{Deng23b}

On the other hand, \textit{prompt-based} approaches are well-suited to directly source instructions from the script and provide a higher degree of control over the behavior of the LLM agent. They allow for explicit selection of dialog acts by the agent as shown in the prompting techniques listed above. Therefore, we decide for an approach injecting our script instructions via \textit{prompting}.

\subsubsection{Section-Level Instructions}

As discussed above, proactive prompting techniques from recent literature involve selection of the next act and generation of the response to the user \textit{in each turn}.

We assume that providing new instructions to the Dialog LLM on each turn would compromise its ability to develop a natural conversation and to react to unforeseen user utterances. We argue that the Dialog LLM needs visibility to a larger context of instructions and some degree of freedom to do the following:

\begin{itemize}
    \item Choose in what order to fulfill the instructions,
    \item skip a part of the instructions that's already completed, 
    \item follow user utterances and temporarily deviate from its instructions, then return to completing them.
\end{itemize}

We therefore propose prompting the Dialog LLM with a set of instructions \textit{larger} than a single-turn act, allowing it to take multiple turns to complete its instructions while possessing the above degrees of freedom. We define the unit of instructions provided to the Dialog LLM at once as a ``section'' of the script, each section consisting of a series of ``tasks'' and representing a \textit{state} the agent can be in. I.e. during runtime, the agent repeatedly moves to a state and gets a section of instructions assigned, then takes one or multiple turns to complete those, and subsequently moves to the next section.

Examples illustrating the need for section-level instructions can be found in the technical appendix of this paper.

\begin{figure}
    \centering
    \includegraphics[width=1\linewidth]{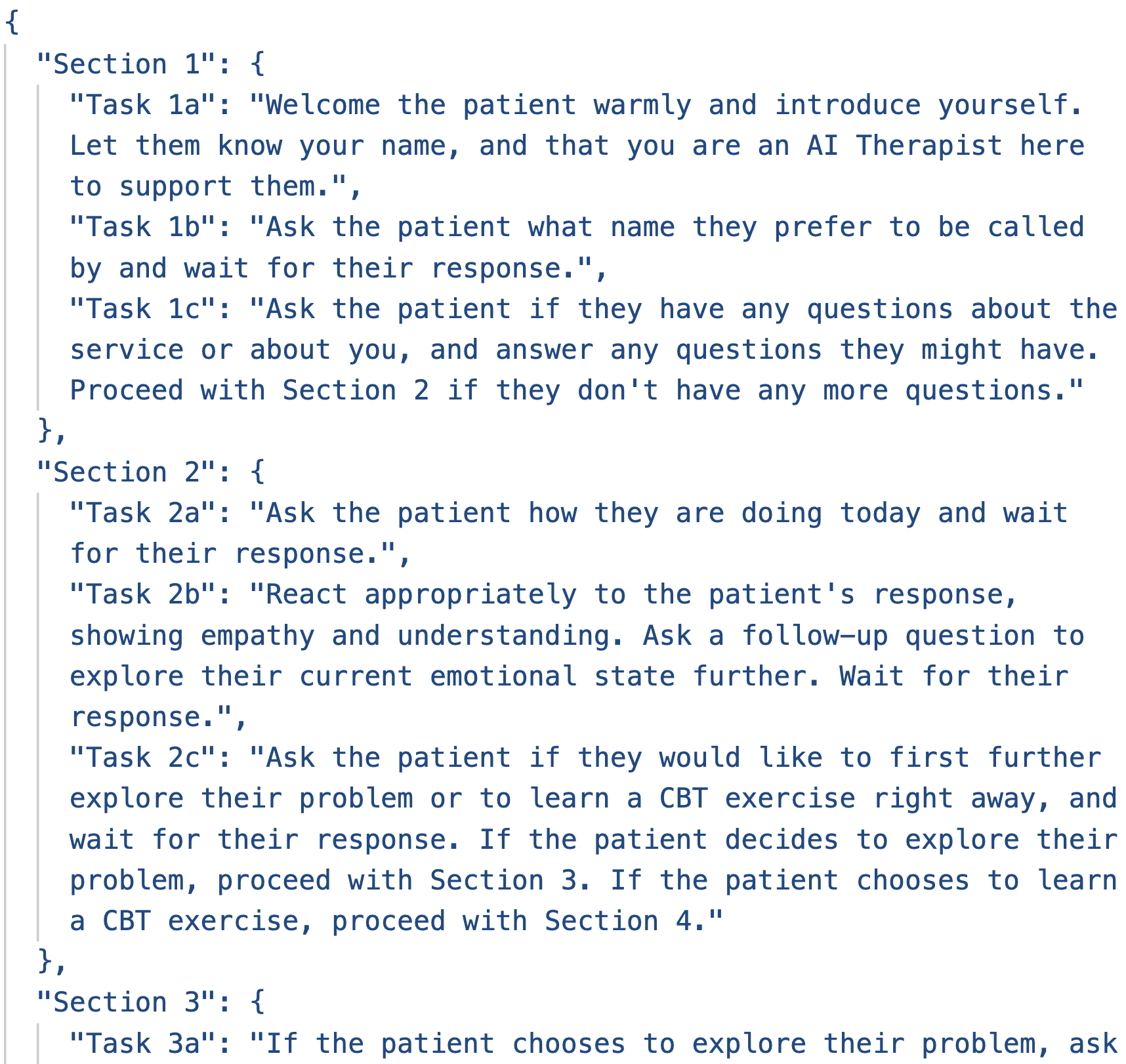}
    \caption{Structure of the exemplary script with multi-task \textit{sections} incl. state transitioning rules (\textit{ ``proceed with ...''}). The complete script is provided in the technical appendix.}
    \label{fig:script-structure}
\end{figure}

An example of a JSON-formatted script structured into multi-task sections in shown in Figure \ref{fig:script-structure}.

\subsubsection{Completion Assessment Step}

The section-level instructions scope requires an additional step on each turn: \textit{Assessment on whether the current instructions have already been completed} by the agent or not, i.e. if the agent should transition to the next state or remain in the current one. Therefore, our Script-Based Dialog Policy Planning involves the following steps performed by the agent on each turn:

\begin{enumerate}
    \item Assessing whether the current instructions have been completed
    \item (Optional \& conditional on instructions completion) Reasoning and planning steps to prepare the decision and response generation
    \item (Conditional on instructions completion) Deciding on the next section of instructions based on the script
    \item Generating the response to the user based on the current section of instructions
\end{enumerate}

Similarly as with proactive prompt-based dialog policy planning, all the steps can be performed either by a \textit{single} LLM, i.e. the Dialog LLM, in a single or in multiple inference runs; or by \textit{multiple} LLM actors which we call Assessor LLM (step 1), Dispatcher LLM (steps 2 \& 3), and Dialog LLM (step 4). In the following section, we will test two implementation variants representing this differentiation.

\begin{figure}
    \centering
    \includegraphics[width=1\linewidth]{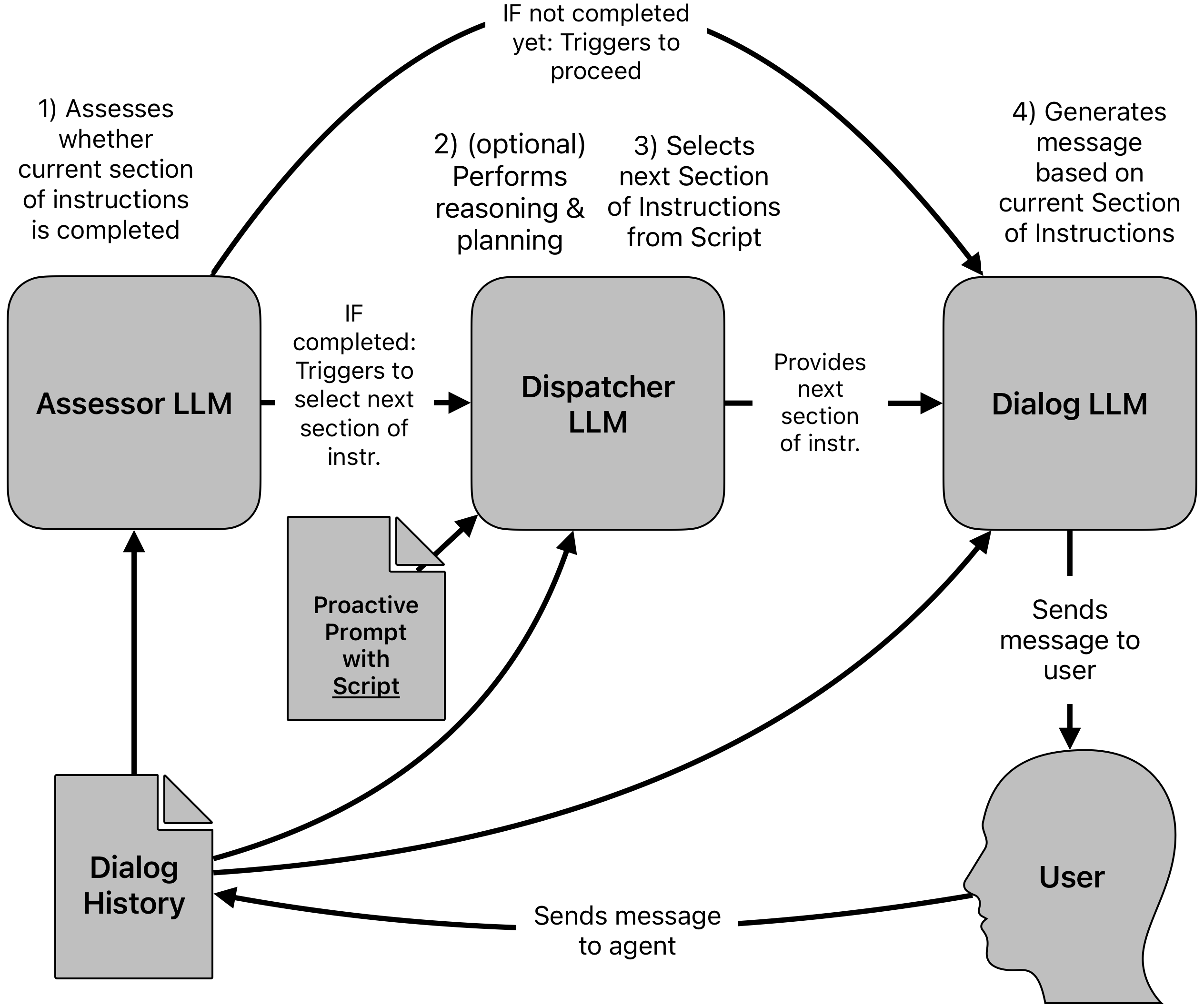}
    \caption{Script-Based Dialog Policy Planning, version with multiple LLM actors: Steps on each conversation turn.}
    \label{fig:script-dpp}
\end{figure}

Figure \ref{fig:script-dpp} presents the process of Script-Based Dialog Policy Planning, showing a version with separate LLM actors.

\section{Experimental Setup}

The objective of this and the next section is to validate the feasibility of our proposed technique. To this end, we have implemented and tested exemplary AI Therapist agents powered by Script-Based Dialog Policy Planning. We have introduced two distinct implementation variants for comparison, with a particular focus on efficiency and effectiveness metrics, in order to gain insights into the optimal implementation of the proposed technique. In order to conduct our experiments, we created synthetic conversations by utilizing an additional LLM to act as a patient. The tests were conducted without the involvement of human users.

Pseudocode of the experiments implementation is included in the paper's technical appendix. The entire software code, that can be used to reproduce and extend the experiments can be found at \href{https://github.com/robderbob/sbdpp}{https://github.com/robderbob/sbdpp}.

\subsection{Exemplary Script}

In our experimental evaluation, we were interested in investigating the \textit{feasibility} of our technique, and not its medical-therapeutic properties (which is future work). Therefore, at this stage, we did not need a real therapeutic manual or a script specifically designed for an AI Therapist.

Instead, the exemplary script needed to be structured according to our defined requirements: Multi-task sections representing states of the conversation, and including instructions on when and how to transition from one state to another. Moreover, the script was chosen to be sufficiently complex to test the transition across its states, but sufficiently short to be able to run a high number of tests within a manageable amount of time and cost.

To fulfill those requirements and still ensure some degree of domain validity, we derived our script from a psychologist-developed chatbot intervention that was part of the study “Corona Stressfrei” (covid stress-free) at the Humboldt-University of Berlin in 2021 \cite{Langhammer21}. Our script matches the high-level conversational flow and phrasing of the above mentioned intervention. The script contains 8 sections (states) with the following contents. State transition rules are shown as ``[$\rightarrow$ \textit{state}]''.

\begin{enumerate}
    \item \textbf{Introduction}: Greet the patient, ask for their name, clarify questions. [$\rightarrow$ 2]
    \item \textbf{Engagement}: Ask the patient for their wellbeing, show empathy, ask if they like to proceed exploring their issue [$\rightarrow$ 3] or conduct a CBT intervention [$\rightarrow$ 4].
    \item \textbf{Exploration}: Ask relevant questions to explore the patient's issue. [$\rightarrow$ 4]
    \item \textbf{Selection}: Based on the patient's condition, suggest a relevant CBT intervention, decide on intervention with the patient. [$\rightarrow$ 5 / 6 / 7]
    \item \textbf{Exercise 1, Thought Record}: Introduce intervention and guide the patient through it. [$\rightarrow$ 8]
    \item \textbf{Exercise 2, Behavioral Activation}: Introduce intervention and guide the patient through it. [$\rightarrow$ 8]
    \item \textbf{Exercise 3, Relaxation Technique}: Introduce intervention and guide the patient through it. [$\rightarrow$ 8]
    \item \textbf{Ending}: Summarize the session, clarify questions, and say goodbye to the patient.
\end{enumerate}

The entire script is included in the technical appendix.

\subsection{Implementation Variants}

Based on the proactive prompting techniques introduced in 2.2 as well as variations discussed in 2.3 with respect to involving \textit{single LLM actor vs. multiple LLM actors}, we chose two different implementation variants for the investigation.

\subsubsection{Variant A: Single LLM Actor}

In variant A, all steps of the dialog policy planning were performed by a \textit{single} ``Dialog LLM'' within a \textit{single} output generation. We used an adaptation of the \textit{ProCoT} prompting scheme \cite{Deng23Procot} and assigned the \textbf{Dialog LLM} to perform the following steps on each turn:

\begin{enumerate}
    \item Assess whether you have completed the current section of instructions and output result as a thought (not visible to the patient).
    \item (IF 1 = completed) Select next section of instructions from the script and output result as a thought (not visible to the patient).
    \item Generate a message to the patient based on the current section of instructions.
\end{enumerate}

In addition to those instructions, the Dialog LLM's system message contains some general acting guidelines as well as the entire script. The Dialog LLM sees the entire dialog history from the \textit{assistant} perspective, i.e. as the actor talking to the user. The Dialog LLM prompt of variant A is included in the technical appendix.

\begin{figure}
    \centering
    \includegraphics[width=1\linewidth]{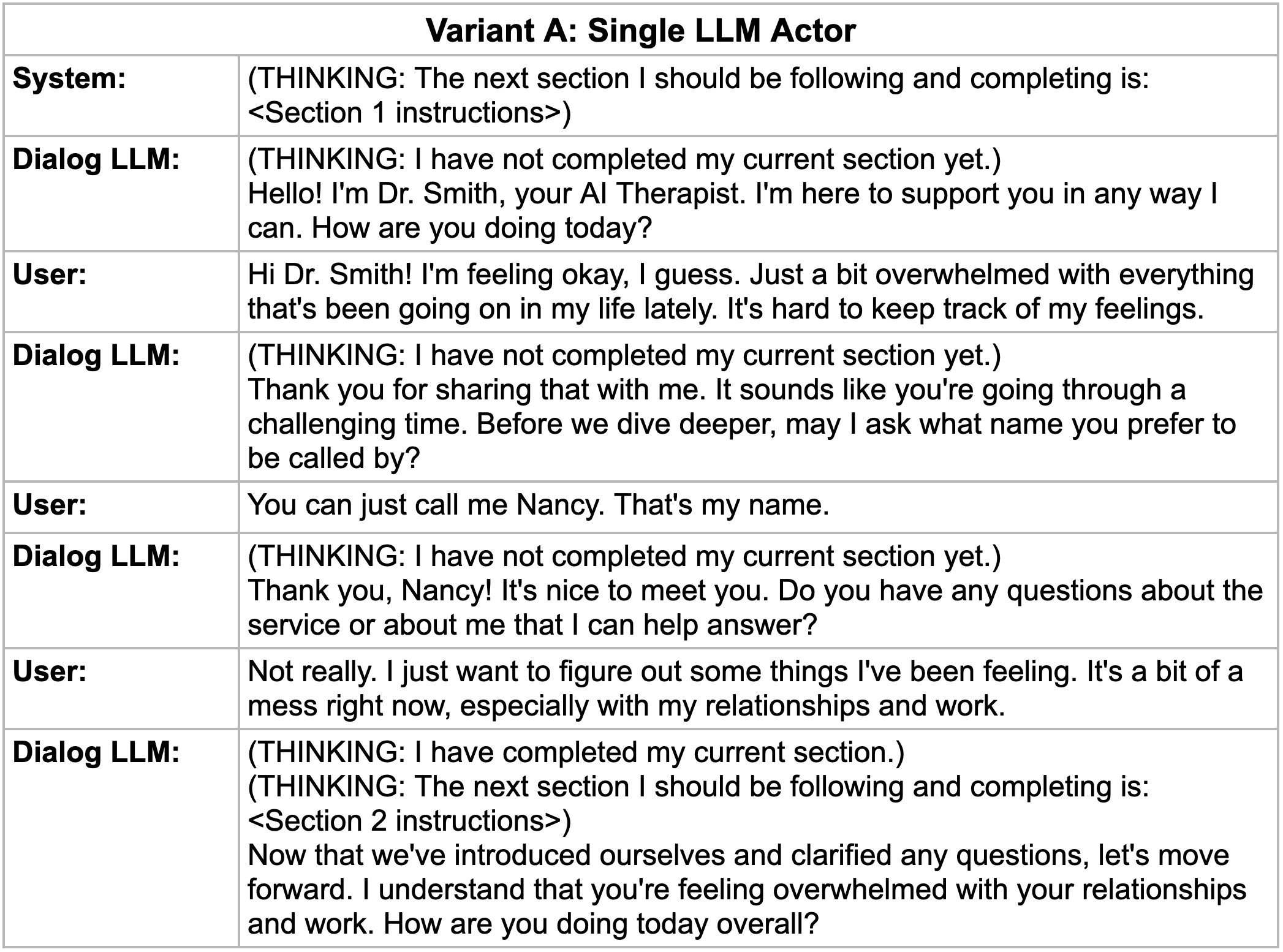}
    \caption{Variant A: Exemplary dialog.}
    \label{fig:variant-a-example}
\end{figure}

Figure \ref{fig:variant-a-example} shows an exemplary dialog demonstrating the behavior of the Dialog LLM in implementation variant A.

\subsubsection{Variant B: Multiple LLM Actors}

In variant B, the steps of the dialog policy planning were performed by \textit{three} different LLM actors within their individual output generations. We were using an adaptation of the \textit{Ask-an-Expert} prompting scheme \cite{Zhang23}, and assigning the LLM actors with their respective tasks on each turn:

\begin{enumerate}
    \item \textbf{Assessor LLM} to assess whether the Dialog LLM has completed the current section of instructions and output result as a thought (not visible to any actor, only used to trigger or skip step 2)
    \item (IF 1 = completed) \textbf{Dispatcher LLM} to select next section of instructions from the script and output result as a thought (only visible to Dialog LLM)
    \item \textbf{Dialog LLM} to generate message to the patient based on the current section of instructions.
\end{enumerate}

In addition to those instructions, the Assessor LLM and the Dispatcher LLM both have access to the current section of instructions as well as the entire dialog history from a \textit{third-party} perspective, i.e. as an ``expert'' watching the therapist-patient conversation. The Dispatcher LLM additionally has access to the entire script. Same as in variant A, the Dialog LLM has acccess to the entire dialog history from the \textit{assistant} perspective. The three LLM actors' prompts of variant B are included in the technical appendix.

\begin{figure}
    \centering
    \includegraphics[width=1\linewidth]{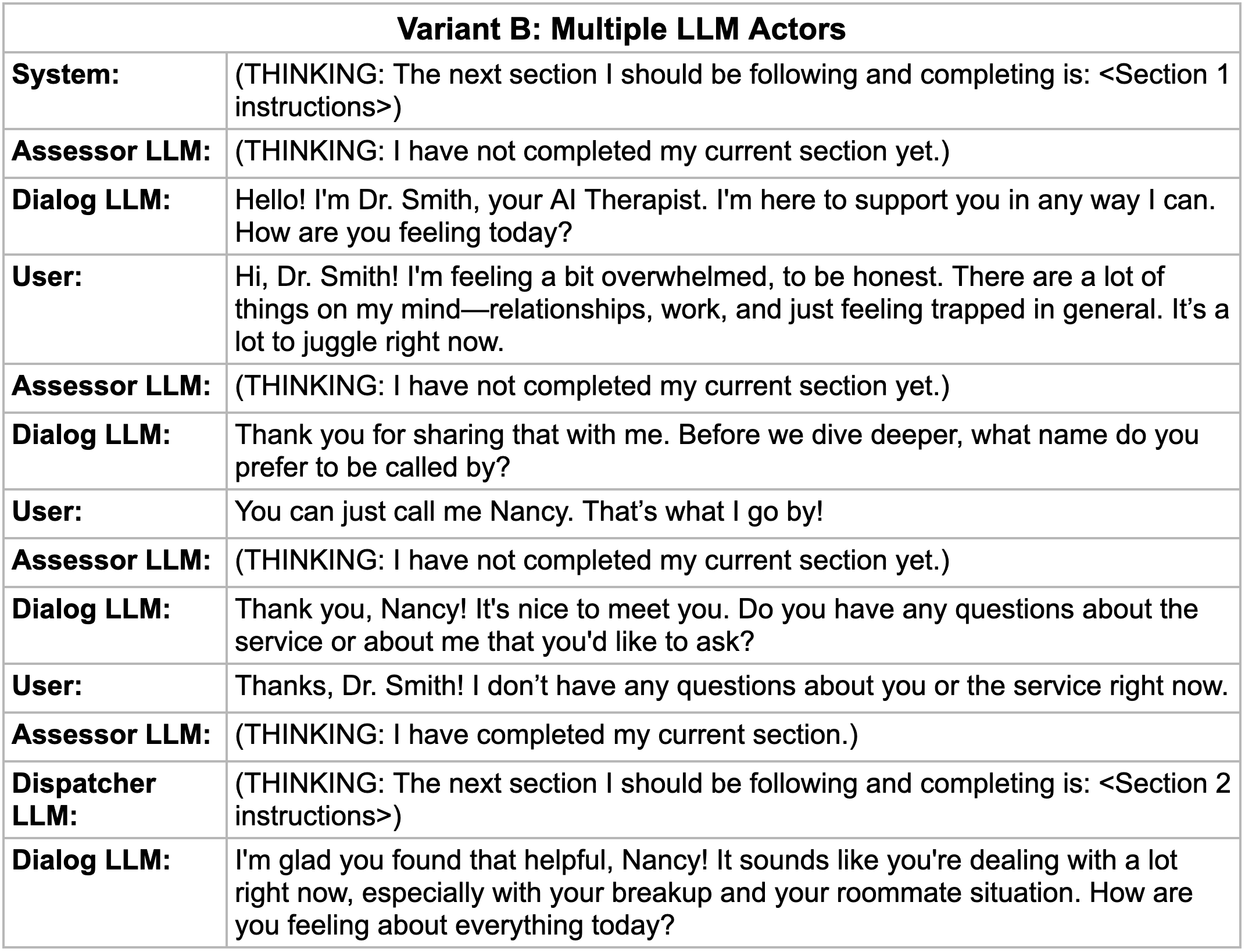}
    \caption{Variant B: Exemplary dialog.}
    \label{fig:variant-b-example}
\end{figure}

Figure \ref{fig:variant-b-example} shows an exemplary dialog demonstrating the involved LLM actors and their interactions in implementation variant B.

\subsection{LLM-Simulated Patients}

In order to perform scaled testing, we have synthesized conversations with LLM-simulated patients. An additional LLM actor was prompted to act like a patient talking to a therapist.

We selected five patient case examples from the American Psychological Association’s Practice Guidelines \cite{apa08}. We used ChatGPT to generate structured summaries of the five patients’ situations, and iteratively fed those into the Patient LLM's prompt for the experiments.

We prompted the Patient LLM to not only act by their case and follow the therapist, but also to ask questions and challenge the therapist's advice from time to time. The patient case summaries and Patient LLM's prompt are included in the technical appendix.

\subsection{Evaluation Metrics}

Proactive conversational agents are usually evaluated on turn- or dialog-level. There are automated as well as human-rated evaluation metrics. This includes metrics like goal completion, user satisfaction, and others \cite{Deng23a}. There are also metrics particularly proposed for \textit{human-centered} PCA \cite{Deng24} and emotional support agents \cite{Zheng23}. However, those are not in the focus of this work. In our evaluation, we are particularly interested in \textit{dialog-level} \textit{efficiency} and \textit{effectiveness} of our agent, and define the relevant metrics as follows.

For \textbf{efficiency}, we are using the following metrics:

\begin{itemize}
    \item \textbf{Avg. Duration per Turn}: Avg. inference time of the involved LLM actors in ms per turn
    \item \textbf{Avg. LLM Input Tokens per Turn}: Avg. input tokens used by the involved LLM actors per turn
    \item \textbf{Avg. LLM Output Tokens per Turn}: Avg. output tokens used by the involved LLM actors per turn
\end{itemize}

In terms of \textbf{effectiveness}, we are using three metrics that represent the LLM actors' performance in our three specific steps of the dialog policy planning. We used:

\begin{itemize}
    \item \textbf{Percentage of Correct Section Completions}: \% of sections that were actually completed when the Dialog LLM (A) resp. the Assessor LLM (B) decided that they are completed --- validated by a Validator LLM.
    \item \textbf{Percentage of Coherent Section Switches}: \% of sections for which the message of the Dialog LLM directly following a section switch (state transition) is \textit{coherent}, i.e. making sense in context of the overall conversation and addressing the latest message of the patient --- validated by a Validator LLM.
    \item \textbf{Percentage of Script-Conform Dispatchings}: \% of sections selected as \textit{next} section by the Dialog LLM (A) resp. the Dispatcher LLM (B) in accordance with the state transitioning rules outlined in the script.
\end{itemize}

The \textbf{Validator LLM} is an additional actor used to validate correct section completions and coherent section switches every time a section is asserted as completed. To this end, the Validator LLM is given the entire dialog history from a \textit{third-party} perspective and prompted to assess the involved LLM actors' performance. The Validator LLM is also asked to \textit{provide reasons} for its decisions. The Validator LLM's prompt is shown in the technical appendix.

We hypothesize that variant A with a single LLM actor is more efficient, whereas variant B with multiple LLM actors is more effective with respect to the above metrics.

\subsection{Used LLMs}

For the Assessor LLM, Dispatcher LLM, and Dialog LLM, we used the LLM \textit{gpt-4o-mini} via the OpenAI API which is considered a state-of-the-art general-use pre-trained LLM with advanced conversational abilities at comparatively low inference duration and cost. We set \textit{temperature=0} to make the behavior as consistent and comparable as possible.

For the Patient LLM, we also used gpt-4o-mini; however, we set \textit{temperature=1} to increase the variability of patient behaviors and allow for more unexpected patient utterances.

For the Validator LLM, we used a more advanced model to allow it to validate the other LLM actors' behavior. To this end, we used \textit{gpt-4o} via the OpenAI API, setting \textit{temperature=0}.

We exclusively used OpenAI LLMs due to their accessibility and low cost. It remains to be shown in later works if the results of this work can be reproduced with other LLMs.

\subsection{Experimental Iterations}

In order to perform our measurements, we have synthesized a total of 100 dialogs: For each of the \textbf{2 implementation variants}, and for each of the \textbf{5 patient cases}, we have generated \textbf{10 dialogs}.

\section{Experimental Results}

\begin{table*}[t]
\centering
{%
\begin{tabular}{llcc}
\toprule
\textbf{Category} & \textbf{Metric} & \textbf{A: Single-LLM} & \textbf{B: Multi-LLM} \\ 
\midrule
\multirow{3}{*}{General Statistics} & \# Total Generated Dialogs & 50 & 50 \\ 
& \# Avg. Turns per Generated Dialog & 22.0 & 23.6 \\ 
& \# Avg. Section Switches per Generated Dialog & 5.84 & 5.82 \\ 
\midrule
\multirow{3}{*}{Efficiency} & Avg. Duration per Turn in ms & \underline{2,800} & 3,520 \\ 
& \# Avg. LLM Input Tokens per Turn & \underline{6,880} & 8,931 \\ 
& \# Avg. LLM Output Tokens per Turn & 151 & \underline{148} \\ 
\midrule
\multirow{3}{*}{Effectiveness} & \% Correct Section Completions & \underline{45.5}\% & 35.7\% \\ 
& \% Coherent Section Switches & \underline{96.9}\% & 87.6\% \\ 
& \% Script-Conform Dispatchings & 86.8\% & \underline{94.2}\% \\ 
\bottomrule
\end{tabular}
}
\caption{Experimental Results for variants A and B. The metric of the winning variant is underlined.}
\label{experimental-results}
\end{table*}

Table \ref{experimental-results} shows the experimental results comparing variants A and B across the selected evaluation metrics based on the 100 generated dialogs.

We have included details on the used hardware and software stack in the technical appendix. The entire dataset of 100 generated dialogs along with computation steps performed to derive the evaluation metrics can be found at \href{https://github.com/robderbob/sbdpp/blob/main/Experiments%20Output%20Data%20and%20Metrics%20Computation.xlsx}{https://github.com/robderbob/sbdpp/blob/main/Exp...}.

\subsubsection{Feasibility}

Reviewing the transcripts of the dialogs, we generally see natural and realistic conversations, the therapist agent leads a coherent discussion with the patient while acting proactively and following the given script. We cannot identify any obvious deficiencies with respect to our outlined requirements. Therefore, we conclude that our proposed architecture is generally \textit{feasible}. Examples of the synthesized dialogs are shown in the technical appendix.

\subsubsection{Efficiency}

Our hypothesis with respect to higher efficiency of variant A was confirmed as variant A shows around 20\% lower avg. duration and 23\% lower avg. input token usage per turn compared to variant B. The difference in avg. output token usage can be neglected.

As assumed, the token usage difference is mainly due to the \textit{single} LLM call per turn in variant A compared to \textit{2-3} calls (depending on the section completion assessment) per turn in variant B. All LLM actors receive the entire dialog history as input, hence increasing the input token usage per LLM call progressively over the course of the dialog. However, input token usage in variant B could be easily reduced by restricting input to only parts of the conversation which are necessary for Assessor LLM and Dispatcher LLM to fulfill their tasks. Moreover, in contrary to variant A, variant B provides the script to the Dispatcher LLM only \textit{selectively} on turns when instructions have been completed. Hence, a longer script would further decrease avg. token usage and duration per turn of variant B relative to variant A.

\subsubsection{Effectiveness}

Variant A shows significantly higher \% correct section completions and \% coherent section switches compared to variant B which is further investigated below.

We mainly see the following and similar reasons provided by the Validator LLM on non-correct section completions: \textit{ ``The therapist did not provide a summary of what was discussed and the exercises conducted during the session.'', ``The therapist did not explicitly ask if the patient had any more questions before proceeding.'', ``The therapist did not wait for the patient's confirmation that the problem summary was correct before proceeding to suggest CBT exercises.''}

Investigating those cases in context of the dialog, we don't find any examples when the LLM actors clearly fail completing their instructions, despite the low \% of correct section completions in both variants. Rather, when the Validator LLM assesses a task completion as non-correct, it is mostly the case that the Dialog LLM is following its instructions \textit{slightly} less accurately for the sake of \textit{addressing a user utterance} and/or \textit{holding a fluent conversation}. Hence, we can't clearly conclude that variant A is completing instructions more correctly, and need to conduct further investigations in future works.

When investigating the \% coherent section switches, we see the following reasons of non-coherence provided by the Validator LLM:  \textit{ ``The therapist did not address the patient's request to talk about what might be contributing to the feeling of being stuck.'', ``The therapist repeated the question ``How are you doing today?'' which was already answered by the patient.'', ``The therapist did not address the patient's question about how they can work together to help her feel less lonely.''}

Looking at the context of the dialogs, we conclude that those cases are more frequent in variant B due to \textit{injection} of new section of instructions by an \textit{external} actor, the Dispatcher LLM, which seems to drag the Dialog LLM towards the newly provided instructions and compromise its ability to accommodate for conversational coherence at section switches.

Finally, we see a higher \% script-conform dispatchings in variant B. When looking at the specific cases, in more than 80\% of non-conform dispatchings, the Dialog LLM resp. Dispatcher LLM decided to move from Section 4 (offering CBT exercises to the patient and asking them to select one) to Section 8 (ending the conversation), even though this is not explicitly allowed by the script. When investigating the conversational context, we almost always find cases where the Patient LLM explicitly stated that ``none of the offered CBT exercises are interesting'' to them or that it ``would like to end the conversation'' for some reasons. We conclude that \textit{in variant A, the Dialog LLM is more inclined to follow the patient's utterances}. This is consistent with our previous observation that \textit{variant B LLM actors are following the instructions more strictly}.

It remains to be evaluated in future works whether it is more desirable to follow the patient's utterance or to strictly adhere to the script, particularly with respect to treatment effectiveness.

\section{Conclusion}

In this work, we introduced a new paradigm for dialog policy planning of conversational agents, called \textit{Script-Based Dialog Policy Planning}, serving the requirements of the behavioral healthcare domain and particularly modeling an AI Therapist. We have outlined the notion of the \textit{script} and explained the associated process that powers the agent's behavior on each turn.

We proposed and implemented two variants of Script-Based Dialog Policy Planning, demonstrating their general feasibility and showing that the variant powered by a single LLM actor is more efficient in terms of token usage, whereas the variant featuring multiple LLM actors is more strictly adhering to its script while compromising following user utterances.

The new architecture requires further development and testing. This includes evaluations with respect to usability and conversational quality of the agent, based on more realistic and extensive expert-developed scripts. Once specific treatment programs have been implemented using the proposed architecture, treatment safety and efficacy must be adequately tested. 

In terms of technical evaluations, future studies should examine different LLMs, prompting techniques, corpus-based vs. prompt-based approaches, and other design variants. Moreover, this work only features automated LLM-based evaluations, and the quality of this technology remains to be validated through interactions with humans.

\section*{Conflict of Interest Statement}

R.W. is co-founder and director of Aury, which develops an AI-based chatbot providing mental health support. K.H. is scientific advisor and received virtual stock options of Aury. C.B. is scientific advisor of Aury.

\bibliography{aaai25}

\end{document}